\documentclass{llncs}
\usepackage{amssymb}
\usepackage{amsmath}

\usepackage{graphicx}
\usepackage[utf8]{inputenc}
\usepackage{color}

\usepackage{graphicx}
\usepackage{tabularx}
\usepackage{xcolor}
\usepackage{float}
\usepackage{rotating}
\usepackage{xstring} 

\usepackage{url}

\newcommand{\R}{\mathbb{R}}
\newcommand{\Rd}{\R^{D}}

\newcommand{\C}{\textbf{\textit{C}}}
\newcommand{\m}{\textbf{\textit{m}}}
\newcommand{\p}{\textbf{\textit{p}}}
\newcommand{\x}{\textbf{\textit{x}}}
\newcommand{\y}{\textbf{\textit{y}}}

\newcommand{\NIPOPa}{NIPOP-aCMA-ES}
\newcommand{\NBIPOPa}{NBIPOP-aCMA-ES}
\newcommand{\NIPOP}{NIPOP-aCMA-ES}
\newcommand{\NBIPOP}{NBIPOP-aCMA-ES}

\newcommand{\ERT}{\ensuremath{\mathrm{ERT}}}

\newcommand{\Df}{\ensuremath{\Delta f}}

\newcommand{\fopt}{\ensuremath{f_\mathrm{opt}}}
\newcommand{\ftarget}{\ensuremath{f_\mathrm{t}}}

\def\mulCMA{$(\mu / \mu_w , \lambda)$-CMA-ES}

\definecolor{mygray}{rgb}{.8,.8,.8}

\begin{document}
\title{Alternative Restart Strategies for CMA-ES}

\author{
Ilya Loshchilov\inst{1}$^{,2}$ 
\and Marc Schoenauer\inst{1}$^{,2}$ \and Mich\`ele Sebag\inst{2}$^{,1}$
}

\institute{
TAO Project-team, INRIA Saclay - \^Ile-de-France\footnote{Work partially funded by FUI of System@tic Paris-Region ICT cluster through contract DGT 117 407 {\em Complex Systems Design Lab} (CSDL). }
\and Laboratoire de Recherche en Informatique 
(UMR CNRS 8623)\\
Universit\'e Paris-Sud, 91128 Orsay Cedex, France\\
{\tt FirstName.LastName@inria.fr}
}

\maketitle

\begin{abstract}

This paper focuses on the restart strategy of CMA-ES on multi-modal functions. A first alternative 
strategy proceeds by decreasing the initial step-size of the mutation while doubling the population
size at each restart. A second strategy adaptively allocates the computational budget among the restart 
settings in the BIPOP scheme. Both restart strategies are validated on the BBOB benchmark; their generality is also demonstrated on an independent real-world problem suite related to spacecraft trajectory optimization.

\end{abstract}

\section{Introduction}

The long tradition of performance of the Covariance Matrix Adaptation Evolution Strategy (CMA-ES) algorithm on real-world problems (with over 100 published applications \cite{CMA-ESApplications}) is due among others to its good behavior on multi-modal functions. Two versions of CMA-ES with restarts have been proposed to handle multi-modal functions: IPOP-CMA-ES \cite{Auger:2005b} was ranked first 
on the continuous optimization benchmark at CEC 2005  \cite{Hansen:2006,CEC2005Garcia2009};
and BIPOP-CMA-ES \cite{DBLP:conf/gecco/Hansen09} showed the best results together 
with IPOP-CMA-ES on the black-box optimization benchmark (BBOB) in 2009 and 2010.

This paper focuses on analyzing and improving the restart strategy of CMA-ES, viewed as 
a noisy hyper-parameter optimization problem in a 2D space (population size, initial step-size).
Two restart strategies are defined. The first one, \NIPOPa\ ({\em New} IPOP-aCMA-ES), differs from IPOP-CMA-ES as it simultaneously
increases the population size and decreases the step size. The second one, \NBIPOPa, allocates computational power to different restart settings depending on their current results. 
While these strategies
have been designed with the BBOB benchmarks in mind \cite{hansen2012fun}, their generality is shown 
on a suite of real-world problems \cite{Vinko2008}. 

The paper is organized as follows. After describing the weighted active $(\mu / \mu_w,\lambda)$-CMA-ES and its current restart strategies (section \ref{section:cma}), the proposed restart schemes are described in section \ref{section:analysisANDalgo}. Section \ref{sectio:experiments} reports on their experimental validation. The paper concludes with a discussion and some perspectives for further research.



\section{The Weighted Active $(\mu / \mu_w,\lambda)$-CMA-ES}
\label{section:cma}
The CMA-ES algorithm is a stochastic optimizer, searching the continuous space $\Rd$ 
by sampling $\lambda$ candidate solutions from a multivariate normal distribution \cite{Hansen:ECJ01,kernhansen:ppsn2004}. 
It exploits the best $\mu$ solutions out of the $\lambda$ ones to adaptively estimate the local 
covariance matrix of the objective function, in order to increase the probability of successful samples in the next iteration. 
The information about the remaining (worst $\lambda-\mu$) solutions is used only implicitly during the selection process.

In active ($\mu/\mu_{I},\lambda$)-CMA-ES however, it has been shown that the worst solutions can be exploited to reduce the variance of the mutation distribution in unpromising directions \cite{activeCMAarnold}, yielding 
a performance gain of a factor 2 for the active ($\mu/\mu_{I},\lambda$)-CMA-ES with no loss of performance 
on any of tested functions.
A recent extension of the \mulCMA, \textit{weighted} active CMA-ES \cite{1830788} (referred to as aCMA-ES for brevity) shows comparable improvements on a set of noiseless and noisy functions from the BBOB 
benchmark suite  \cite{hansen2012exp}. In counterpart, aCMA-ES no longer guarantees the covariance matrix to be positive definite, possibly resulting in algorithmic instability. The instability
issues can however be numerically controlled during the search; as a matter of fact they are never observed on the BBOB benchmark suite.

At iteration $t$, \mulCMA\ samples $\lambda$ individuals according to

\begin{equation}
\label{primal}
\x^{(t+1)}_k \sim
{{\mathcal N}\hspace{-0.13em}\left(\m^{(t)}, {\sigma^{(t)}}^2 \C^{(t)} \right)}, \;\;\; k=1\ldots \lambda, 
\end{equation}

where ${\mathcal N}\hspace{-0.13em}\left(\m,\C \right)$ denotes a normally distributed random vector with mean \m\ and covariance matrix \C.

These $\lambda$ individuals are evaluated and ranked, where  index $i:\lambda$ denotes the $i$-th best individual after the objective function. The mean of the distribution is updated and set to 
the weighted sum of the best $\mu$ individuals ($\m = \sum^{\mu}_{i=1} w_i \x^{(t)}_{i:\lambda}$, 
with $w_i>0$ for $i=1\ldots \mu$ and $\sum^{\mu}_{i=1} w_i = 1$).

The active CMA-ES only differs from the original CMA-ES in the adaptation of the covariance matrix $\C^{(t)}$.
Like for CMA-ES, the covariance matrix is computed from the best $\mu$ solutions, 
$ \C^{+}_{\mu} = \sum^{\mu}_{i=1} w_i \frac{\x_{i:\lambda} - \m^t}{\sigma^t} \times \frac{(\x_{i:\lambda} - \m^t)^T}{\sigma^t}$. The main novelty is to exploit the worst solutions to compute 
$ \C^{-}_{\mu} = \sum^{\mu-1}_{i=0} w_{i+1} \y_{\lambda-i:\lambda}  \y^T_{\lambda-i:\lambda}$, where 
$ \y_{\lambda-i:\lambda} = \frac{\left\|  \ {C^{t}}^{-1/2} (\x_{\lambda-\mu+1+i:\lambda}-\m^t) \right\|}{ \left\|  \ {C^{t}}^{-1/2} (\x_{\lambda-i:\lambda}-\m^t) \right\| }
\times \frac{\x_{\lambda - i:\lambda} - \m^t}{\sigma^t}$. The covariance matrix estimation of these worst solutions is used to decrease the variance of the mutation distribution along these directions:

\begin{equation}
\begin{array}{rl}
\lefteqn{ \C^{t+1} = (1 - c_1 - c_{\mu} + c^{-}\alpha^{-}_{old}) \C^t + }\\
& + c_1 \p^{t+1}_c {\p^{t+1}_c}^T + (c_{\mu} + c^{-}(1-\alpha^{-}_{old}))\C^{+}_{\mu} - c^{-}\C^{-}_{\mu},
\end{array}
\end{equation}

where $\p^{t+1}_c$ is adapted along the evolution path and coefficients $c_1$, $c_{\mu}$, $c^{-}$ and $\alpha^{-}_{old}$ are defined such that $c_1 + c_{\mu} - c^{-}\alpha^{-}_{old} \leq 1$. The 
interested reader is referred to \cite{Hansen:ECJ01,1830788} for 
a more detailed description of these algorithms.
%

As mentioned, CMA-ES has been extended with restart strategies to accommodate multi-modal fitness landscapes, and to specifically handle objective functions with many local optima. 
As observed by  \cite{kernhansen:ppsn2004}, the probability of reaching the optimum (and the overall number of function evaluations needed to do so) is very sensitive to the population size. 
The default population size $\lambda_{default}$ has been tuned for uni-modal functions; it is hardly large enough for multi-modal functions. Accordingly, \cite{Auger:2005b} proposed a ``doubling trick'' restart strategy
to enforce global search: the restart \mulCMA\ with increasing population, called IPOP-CMA-ES, 
is a multi-restart strategy where the population size of the run is doubled in each restart until meeting a stopping criterion. 
%


The BIPOP-CMA-ES instead considers two restart regimes. The first one, which corresponds to IPOP-CMA-ES,
doubles the population size $\lambda_{large}=2^{i_{restart}}\lambda_{default}$ 
in each restart $i_{restart}$ and uses a fixed initial step-size $\sigma^0_{large}=\sigma^0_{default}$.\\
The second regime uses a small population size $\lambda_{small}$ and initial step-size $\sigma^0_{small}$,
which are randomly drawn in each restart as:
\begin{equation}
\begin{array}{rl}
\lambda_{small} = \left\lfloor \lambda_{default} \left( \frac{1}{2} \frac{\lambda_{large}}{\lambda_{default}} \right)^{U[0,1]^2}   \right\rfloor, \hspace*{.1in} \sigma^0_{small}=\sigma^0_{default} \times 10^{-2 U[0,1]}
\end{array}
\end{equation}
where $U[0,1]$ stands for the uniform distribution in $[0,1]$. Population size $\lambda_{small}$ thus varies  $\in [\lambda_{default}, \lambda_{large}/2]$.
BIPOP-CMA-ES launches the first run with default population size and initial step-size. In each restart, 
it selects the restart regime with less function evaluations. Clearly, the second regime consumes less 
function evaluations than the doubling regime; it is therefore launched more often.

%
%
%
%
%
%

\section{Alternative Restart Strategies}
\label{section:analysisANDalgo}
\subsection{Preliminary Analysis}

The restart strategies of IPOP- and BIPOP-CMA-ES are viewed as a search in the hyper-parameter space.

\begin{figure*}[tb]
\begin{center}
\begin{tabular}{cc}
  \includegraphics[scale=0.47]{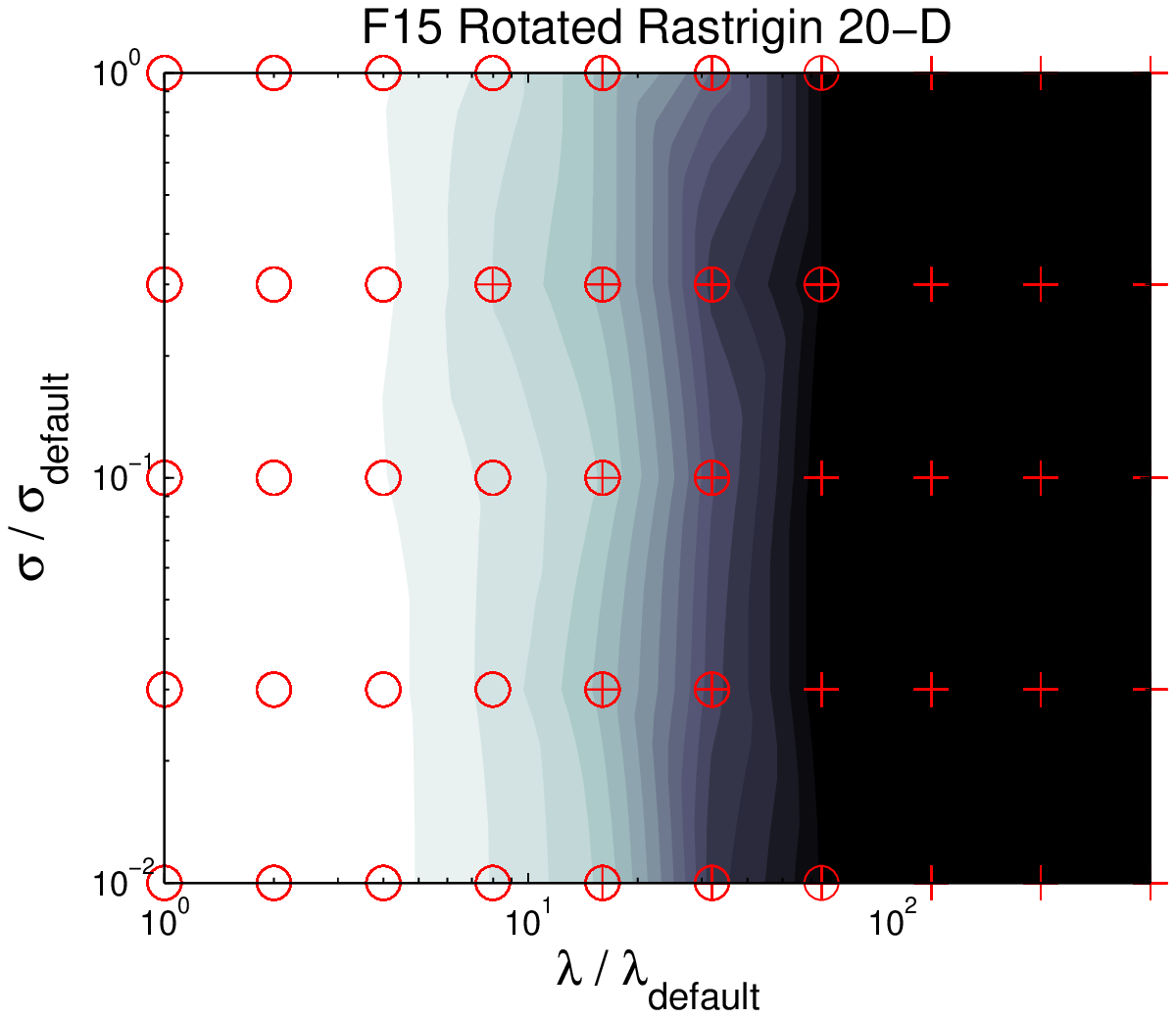}
	\includegraphics[scale=0.47]{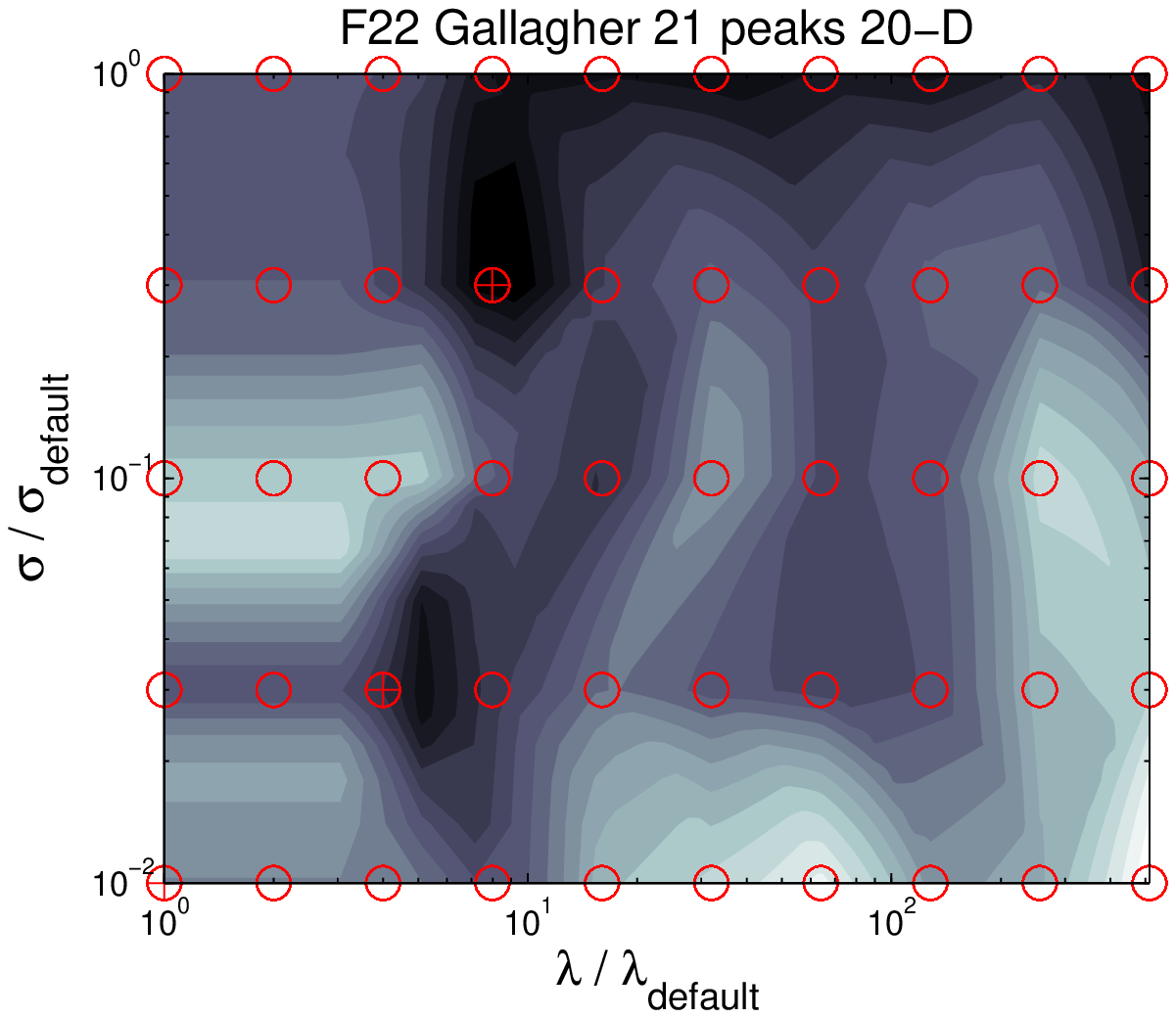}\\
	\includegraphics[scale=0.47]{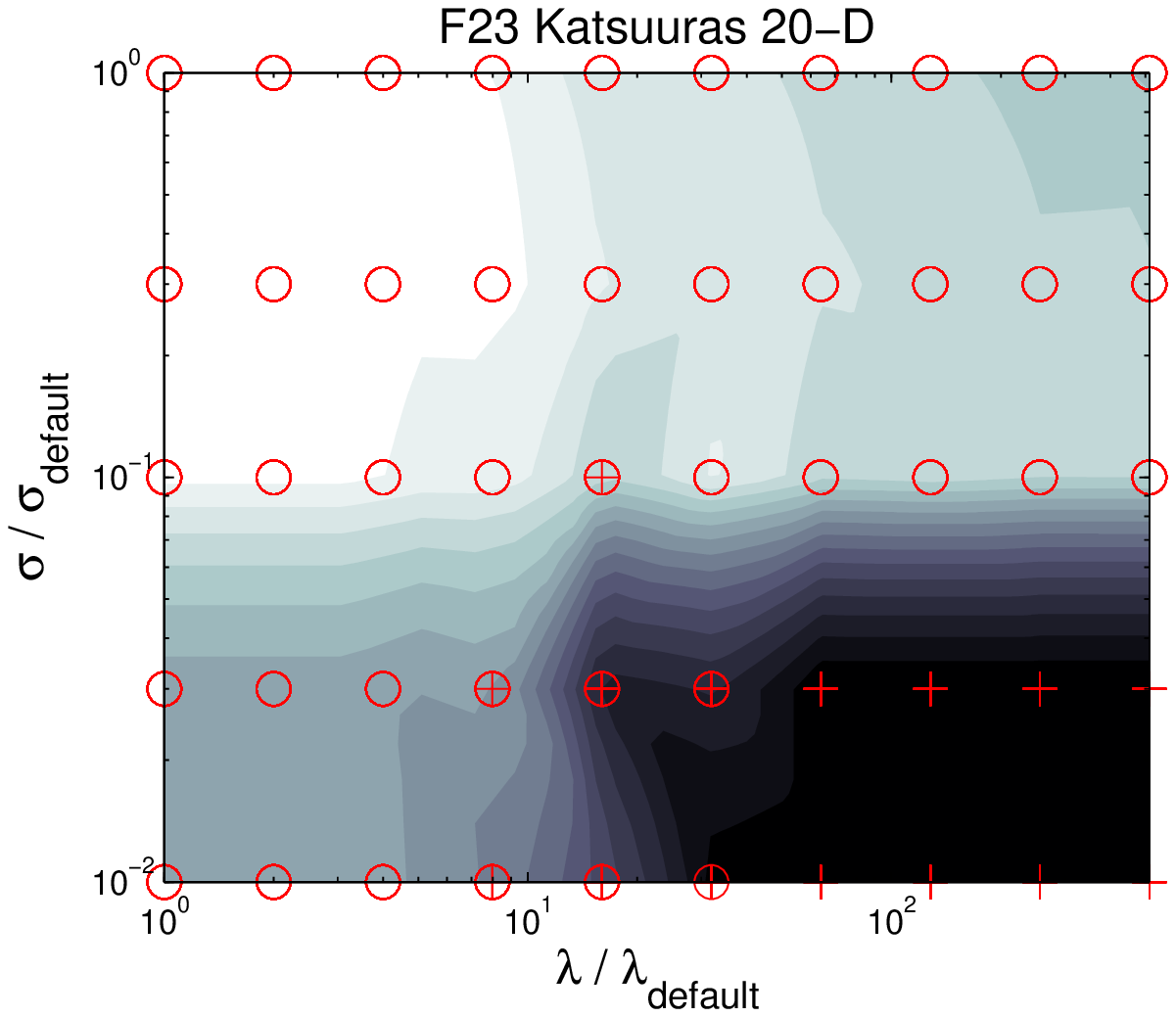}
  \includegraphics[scale=0.47]{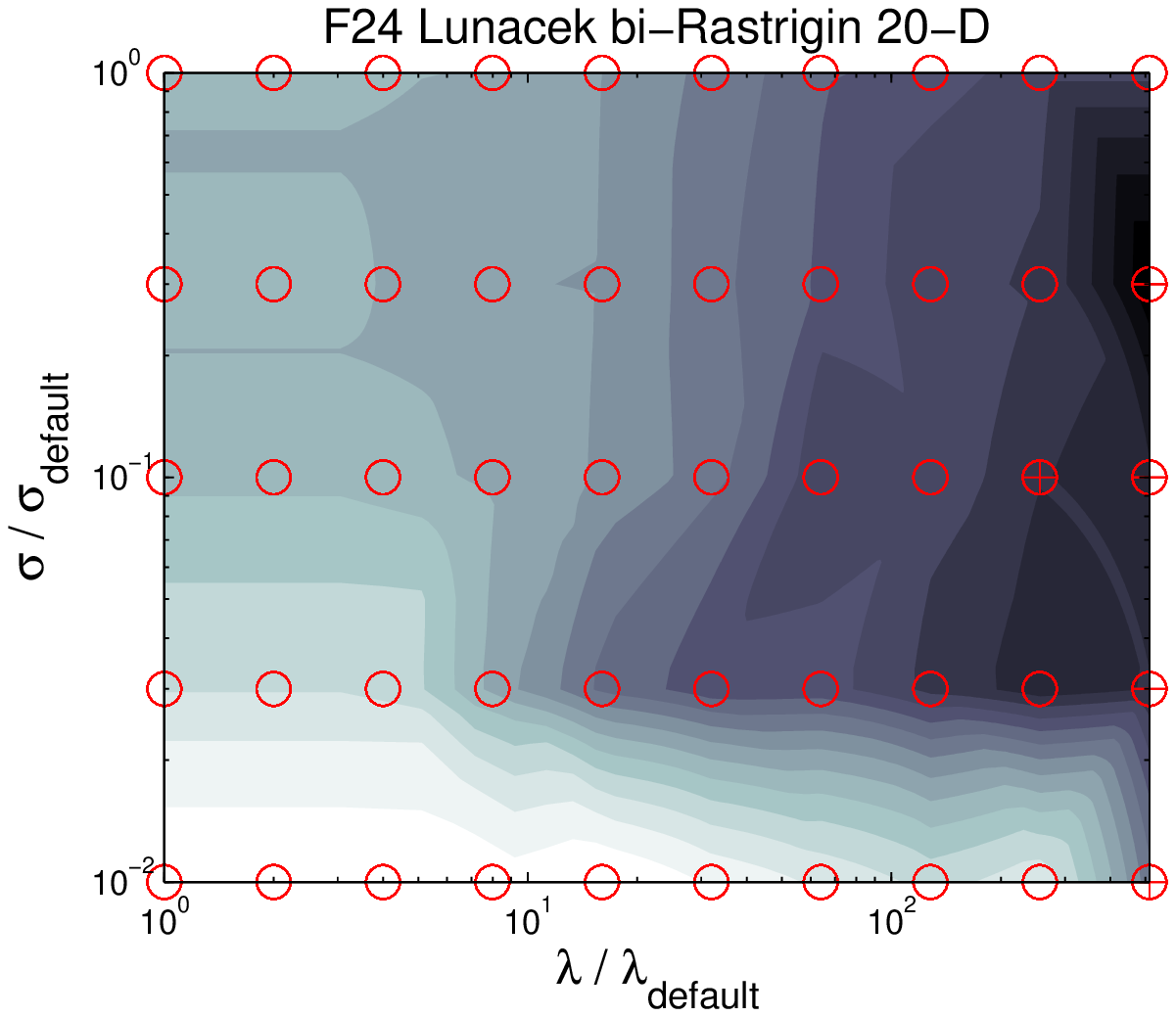}
\end{tabular}
\end{center}
\caption{\label{fig:lambdasigma} Restart performances in the 2D hyper-parameter space (population size and initial mutation step size in log. coordinates). For each objective function (20 dimensional Rastrigin - top-left, Gallagher 21 peaks - top-right, Katsuuras - bottom-left and Lunacek bi-Rastrigin bottom-right), the median best function value out of 15 runs
is indicated. 
Legends indicate that the optimum up to precision $f(x)=10^{-10}$ is found always ($+$), sometimes ($\oplus$) or never ($\circ$).
Black regions are better than white ones.}
\end{figure*}

IPOP-CMA-ES only aims at adjusting population size $\lambda$. It is motivated by the results  observed on multi-modal problems \cite{kernhansen:ppsn2004}, suggesting that the population size must be sufficiently large to 
handle problems with global structure. In such cases, a large population size is needed to uncover this global structure and to lead the algorithm to discover the global optimum. 
IPOP-CMA-ES thus increases the population size in each restart, irrespective of the results observed 
so far; at each restart, it launches a new CMA-ES with population size $\lambda=\rho_{inc}^{i_{restart}}\lambda_{default}$ (see $\circ$ on Fig. \ref{fig:lambdasigma}). 
Factor $\rho_{inc}$ must be not too large to avoid "overjumping" some possibly optimal population size $\lambda^{\ast}$; it must also be not too small in order to reach $\lambda^{\ast}$ in a reasonable number of restarts. The use of the doubling trick ($\rho_{inc}=2$) guarantees that the loss in terms of function evaluations (compared to the ``oracle`` restart strategy which would directly set the population size to the optimal value $\lambda^{\ast}$) is about a factor of 2.

On the Rastrigin 20-D function, IPOP-CMA-ES performs well and always finds the optimum after about 5 restarts (Fig. \ref{fig:lambdasigma}, top-left). The Rastrigin function displays indeed a global structure  where the optimum is the minimizer of this structure.
For such functions, IPOP-CMA-ES certainly is the method of choice. For some other functions such as the 
Gallagher function, there is no such global structure; increasing the population size does not improve the results. 
On Katsuuras and Lunacek bi-Rastrigin functions, the optimum can only be found with small initial step-size (lesser than the default one); this explains why it can be solved by BIPOP-CMA-ES, sampling the two-dimensional ($\lambda, \sigma$) space.

Actually, the optimization of a multi-modal function by CMA-ES with restarts can be viewed as the optimization of the function $h(\theta)$, 
which returns the optimum found by CMA-ES defined by the hyper-parameters $\theta$=($\lambda,\sigma$). 
Function $h(\theta)$, graphically depicted in Fig. \ref{fig:lambdasigma} can be viewed as a black box, computationally expensive
and stochastic function (reflecting the stochasticity of CMA-ES).
Both IPOP-CMA-ES and BIPOP-CMA-ES are based on implicit assumptions about the $h(\theta)$: IPOP-CMA-ES achieves a deterministic uni-dimensional
trajectory, and BIPOP-CMA-ES randomly samples the 2-dimensional search space.

Function $h(\theta)$ also can be viewed as a multi-objective fitness, since in addition to the solution found by CMA-ES, 
$h(\theta)$ could return the number of function evaluations needed to find that solution. 
$h(\theta)$ could also return the computational effort SP1 
(i.e. the average number of function evaluations of all successful runs, divided by proportion of successful runs). 
However, SP1 can only be known for benchmark problems where
the optimum is known; as the empirical optimum is used in lieu of true optimum, SP1 can only be computed {\em a posteriori}.

\subsection{Algorithm}

\begin{figure*}[tb]
\begin{center}
  \includegraphics[scale=0.60]{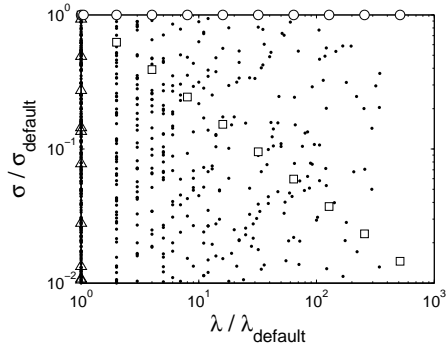}
\end{center}
\caption{\label{fig:algorithm} An illustration of $\lambda$ and $\sigma$ hyper-parameters distribution 
for 9 restarts of IPOP-aCMA-ES ($\circ$), BIPOP-aCMA-ES ($\circ$ and $\cdot$ for 10 runs),
\NIPOPa\ ($\square$) and \NBIPOPa\ ($\square$ and many $\bigtriangleup$ for $\lambda/\lambda_{default}=1$, $\sigma/\sigma_{default} \in [10^{-2},10^0]$).
The first run of all algorithms corresponds to the point with $\lambda/\lambda_{default}=1$, $\sigma/\sigma_{default}=1$.}
\end{figure*}

Two new restart strategies for CMA-ES, respectively referred to as \NIPOP\ and \NBIPOP, are presented in this paper.

If the restart strategy is restricted to the case of increasing of population size (IPOP), 
we propose to use \NIPOP, where we additionally decrease the initial step-size by some factor $\rho_{\sigma dec}$.
The rationale behind this approach is that the CMA-ES with relatively small initial step-size
is able to explore small basins of attraction (see Katsuuras and Lunacek bi-Rastrigin functions on Fig. \ref{fig:lambdasigma}), 
while with initially large step-size and population size 
it will neglect the local structure of the function, but converge to the minimizer of the global structure.
Moreover, initially, relatively small step-size will quickly increase if it makes sense, 
and this will allow the algorithm to recover the same global search properties than 
with initially large step-size (see Rastrigin function on Fig. \ref{fig:lambdasigma}).

NIPOP-CMA-ES thus explores the two-dimensional hyper-parameter space in a deterministic way (see $\square$ symbols on Fig. \ref{fig:algorithm}).
For $\rho_{\sigma dec}=1.6$ used in this study, NIPOP-CMA-ES thus reaches the lower bound ($\sigma=10^{-2} \sigma_{default}$) used by BIPOP-CMA-ES after 9 restarts, expectedly reaching the same performance as BIPOP-CMA-ES albeit it uses only a large population.

The second restart strategy, \NBIPOP, addresses the case where the probability to find the global optimum does not much vary in the $(\lambda,\sigma)$ space. 
Under this assumption, it makes sense to have many restarts for a fixed budget (number of function evaluations). 
Specifically, \NBIPOP\ implements the competition of the \NIPOP\ strategy (increasing $\lambda$ and decreasing initial $\sigma^0$ in each restart) 
and a uniform sampling of the $\sigma$ space, where $\lambda$ is set to $\lambda_{default}$ and $\sigma^0=\sigma^0_{default} \times 10^{-2 U[0,1]}$
The selection between the two (\NIPOP\ and the uniform sampling) depends on the allowed budget
like in \NBIPOP. 
The difference is that \NBIPOP\ adaptively sets the budget allowed to each
restart strategy, where the restart strategy leading to the overall best solution found so far is allowed twice ($\rho_{budget}=2$) a budget compared to the other strategy.


\section{Experimental Validation}\label{sectio:experiments}
The experimental validation of \NIPOPa\ and \NBIPOPa\ investigates the performance of the
approach comparatively to IPOP-aCMA-ES and BIPOP-aCMA-ES on BBOB noiseless problems and 
one black-box real-world problem related to spacecraft trajectory optimization.
The default parameters of CMA-ES \cite{1830788,DBLP:conf/gecco/Hansen09} are used.
This section also presents the first experimental study of BIPOP-aCMA-ES\footnote{For the sake of reproducibility, the source code for \NIPOPa\ and \NBIPOPa\ is available at \url{https://sites.google.com/site/ppsnbipop/}}, the active version of BIPOP-CMA-ES \cite{DBLP:conf/gecco/Hansen09}.

\subsection{Benchmarking with BBOB Framework}

The BBOB framework \cite{hansen2012exp} is made of 24 noiseless and 30 noisy functions \cite{hansen2012fun}. Only the noiseless case has been considered here.
Furthermore, only the 12 multi-modal functions among these 24 noiseless functions are of interest for this study, as CMA-ES can solve the 12 other functions without any restart.

With same experimental methodology as in \cite{hansen2012exp}, the results obtained on these benchmark functions
are presented in Fig. \ref{fig:ECDFs40D} and Table \ref{tab:ERTs40}. 
The results are given for dimension $40$, because the differences are larger in higher dimensions. 
The \textbf{expected running time (ERT)}, used in the figures and table,
depends on a given target function value, $\ftarget=\fopt+\Df$. It is
computed over all relevant trials as the number of function
evaluations required in order to reach \ftarget, summed over all 15 trials, and divided by the
number of trials that actually reached \ftarget\
\cite{hansen2012exp}. 

\textbf{\NIPOPa}. On 6 out of 12 test functions ($f_{15}$,$f_{16}$,$f_{17}$,$f_{18}$,$f_{23}$,$f_{24}$) \NIPOPa\ obtains the best known results for BBOB-2009 and BBOB-2010 workshops.
On $f_{23}$ Katsuuras and $f_{24}$ Lunacek bi-Rastrigin, \NIPOPa\ has a speedup of a factor from 2 to 3, as could have been expected.
It performs unexpectedly well on $f_{16}$ Weierstrass functions, 7 times faster than IPOP-aCMA-ES and almost 3 times faster than BIPOP-aCMA-ES.
Overall, according to Fig. \ref{fig:ECDFs40D}, \NIPOPa\ performs as well as BIPOP-aCMA-ES, while restricted to only one regime of increasing population size.

\textbf{\NBIPOPa}. Thanks to the first regime of increasing population size, \NBIPOPa\ inherits some results of \NIPOPa.
However, on functions where the population size does not play any important role, it performs significantly better than BIPOP-aCMA-ES.
This is the case for $f_{21}$ Gallagher 101 peaks and $f_{22}$ Gallagher 21 peaks functions, where \NBIPOPa\ has a speedup of a factor of 6.
It seems that the adaptive choice between two regimes works efficiently on all functions except on $f_{16}$ Weierstrass. In this last case,
\NBIPOPa\ mistakingly prefers small populations, with a loss factor 4 compared to \NIPOPa.
According to Fig. \ref{fig:ECDFs40D}, \NBIPOPa\ performs better than BIPOP-aCMA-ES on weakly structured multi-modal functions,
showing overall best results for BBOB-2009 and BBOB-2010 workshops in dimensions 20 (results not shown here) and 40.

Due to space limitations, the interested reader is referred to \cite{BBOB2012NIPOP} for a detailed presentation of the
results.

\subsection{Interplanetary Trajectory Optimization}

The \NIPOPa\ and \NBIPOPa\ strategies, designed for the BBOB benchmark functions, can possibly {\em overfit}
this benchmark suite. In order to test the generality of these strategies, a real-world black-box problem
is considered, pertaining to a completely
different domain: Advanced Concepts Team of European Space Agency is making
available several difficult spacecraft trajectory optimization problems as black
box functions to invite the operational research community to compare different
derivative-free solvers on these test problems \cite{Vinko2008}.

The following results consider the 18-dimensional bound-constrained black-
box function "TandEM-Atlas501", that defines an interplanetary trajectory to
Saturn from the Earth with multiple fly-bys, launched by the rocket Atlas 501.
The final goal is to maximize the mass $f(x)$, which can be delivered to Saturn using one of 24 possible fly-by sequences
with possible maneuvers around Venus, Mars and Jupiter.

The first best results was found for a sequence Earth-Venus-Earth-Earth-Saturn ($f_{max}=1533.45$) in 2008 by B. Addis et al. \cite{Addis2008}. 
The best results so far ($f_{max}=1673.88$) was found in 2011 by G. Stracquadanio et al. \cite{Stracquadanio2011}. 

All versions of CMA-ES with restarts have been launched with a maximum budget of $10^8$ function evaluations.
All variables are normalized in the range $[0,1]$.
In the case of sampling outside of boundaries, the fitness is penalized and becomes $f(x) = f(x_{feasible}) - \alpha \left\| x - x_{feasible} \right\|^2$, 
where $x_{feasible}$ is the closest feasible point from point $x$ and $\alpha$ is a penalty factor, which was arbitrarily set to $1000$.

\begin{figure*}[tb]
\begin{center}
\begin{tabular}{cc}
  \includegraphics[scale=0.45]{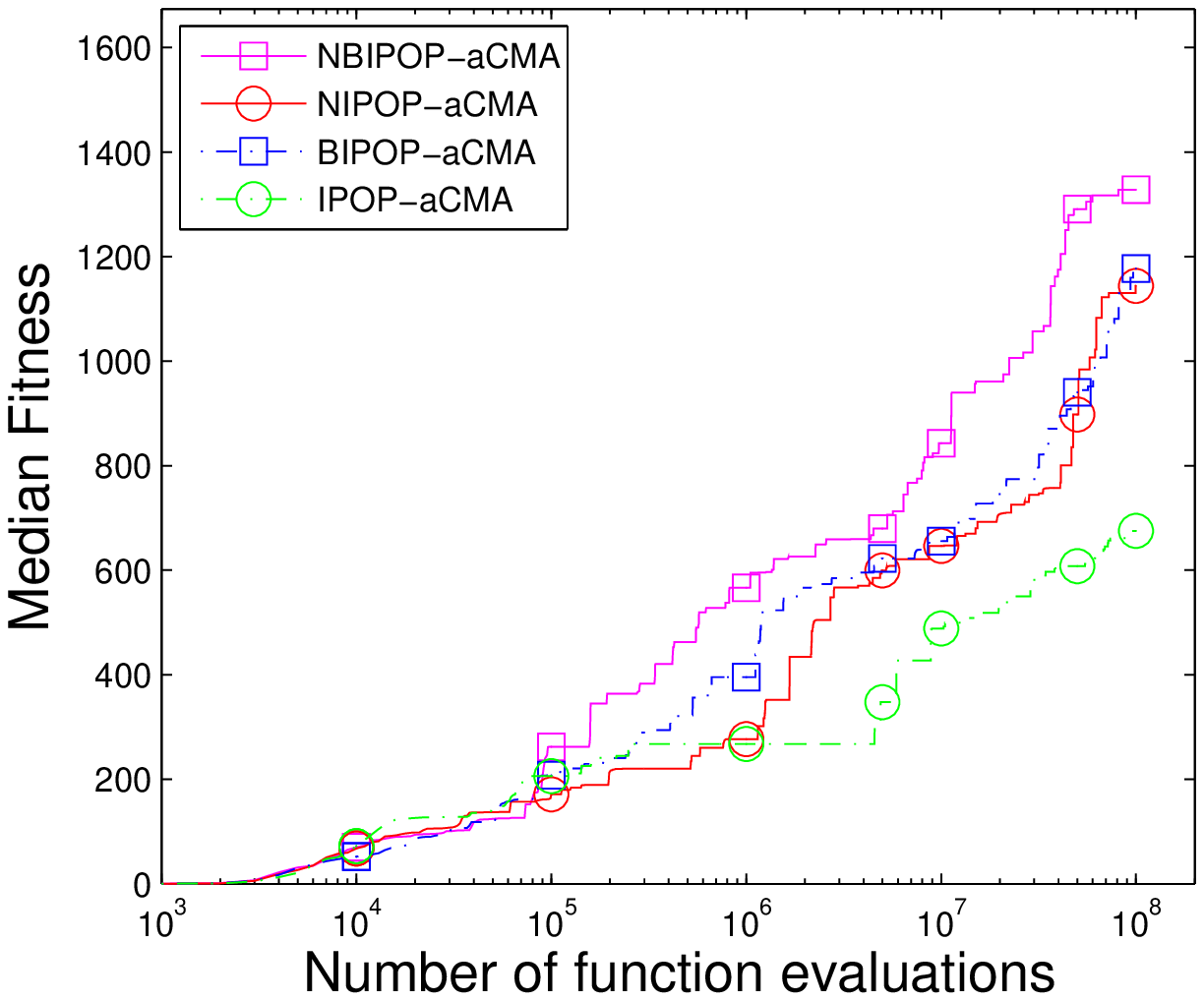}
  \includegraphics[scale=0.45]{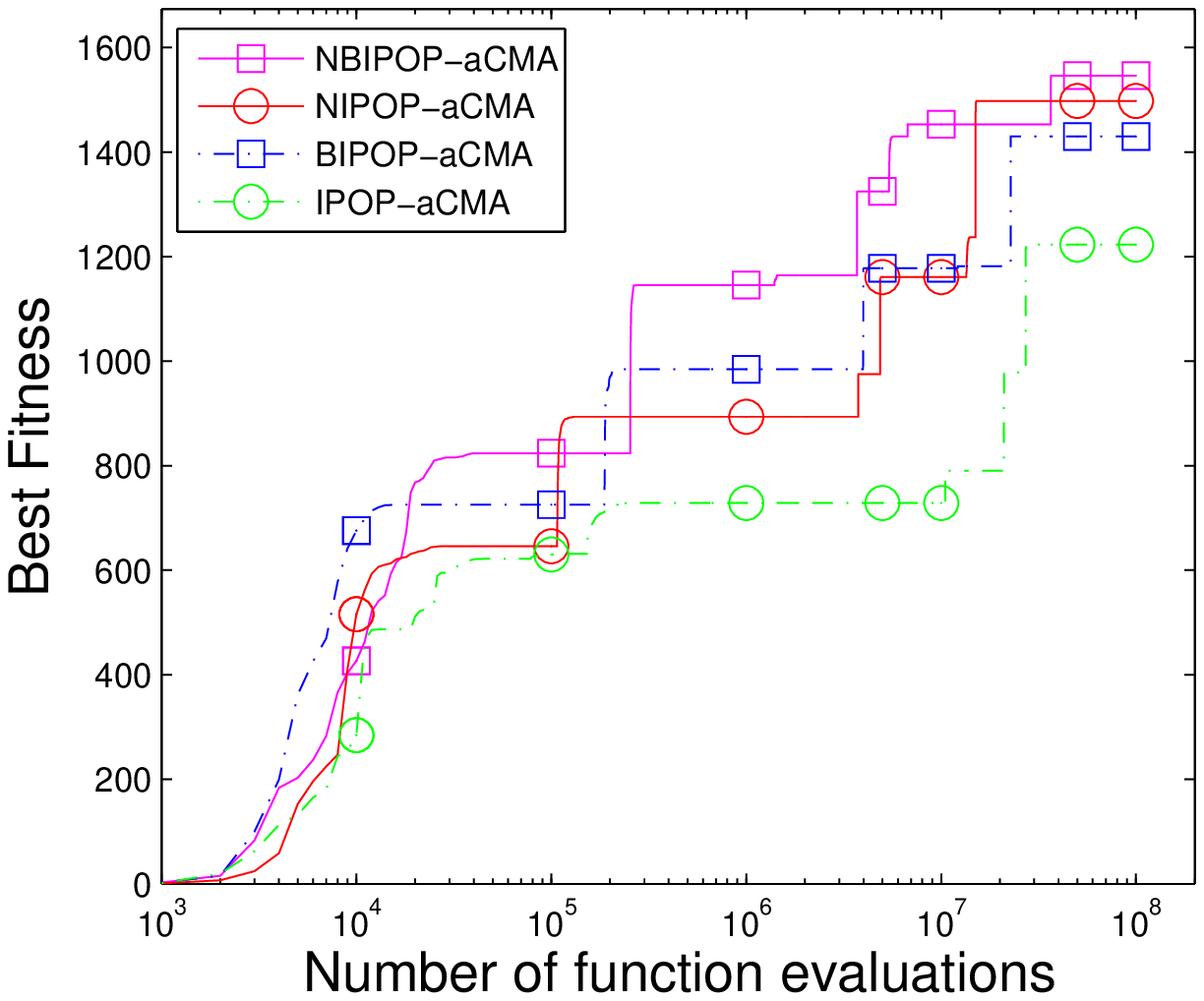}	
\end{tabular}
\end{center}
\caption{\label{fig:tandem} Comparison of all CMA-ES restart strategies on the Tandem fitness function (mass): median (left) and best (right)
values out of 30 runs.}
\end{figure*}

As shown on Fig. \ref{fig:tandem}, the new restart strategies \NIPOPa\ and \NBIPOPa\ respectively improve on the
former ones (IPOP-aCMA-ES and BIPOP-aCMA-ES); further, \NIPOPa\ reaches same performances as BIPOP-aCMA-ES.

The best solution found by \NBIPOPa\ \footnote{ $x=$[0.83521, 0.45092, 0.50284, 0.65291, 0.61389, 0.75773, 0.43376, 1, 0.89512, 0.77264, 0.11229, 0.20774, 0.018255, 6.2057e-09, 4.0371e-08, 0.2028, 0.36272, 0.32442]; fitness(x) = mass(x) = 1546.5}
improves on the best solution found in 2008, while it is worse than the current best solution,
which is blamed on the lack of problem specific heuristics \cite{Addis2008,Stracquadanio2011}, on the possibly insufficient time
budget ($10^{8}$ fitness evaluations), and also on the lack of appropriate constraint handling heuristics.

\section{Conclusion and Perspectives}
\label{section:conclusion}

This paper contribution regards two new restart strategies for CMA-ES. \NIPOPa\
is a deterministic strategy simultaneously increasing the population size {\em and}
decreasing the initial step-size of the Gaussian mutation. \NBIPOPa\ implements a competition
between \NIPOPa\ and a random sampling of the initial mutation step-size, adaptively adjusting
the computational budget of each one depending on their current best results.
Besides the extensive validation of \NIPOPa\ and \NBIPOPa\ on the BBOB benchmark, the generality of these strategies has been tested on a new problem, related to interplanetary
spacecraft trajectory planning.

The main limitation of the proposed restart strategies is to quasi implement a deterministic
trajectory in the $\theta$ space. Further work will consider $h(\theta)$ as yet another expensive noisy black-box function, and the use of a CMA-ES in the hyper-parameter space will be studied. The critical issue is naturally to
keep the overall number of fitness evaluations beyond reasonable limits. A surrogate-based approach
will be investigated \cite{ACMGECCO2012}, learning and exploiting an estimate of the (noisy and stochastic) $h(\theta)$
function.

\bibliographystyle{abbrv}

 \newcommand{\algaperfprof}{BIPOP-aCMA}
 \newcommand{\algbperfprof}{IPOP-aCMA}
 \newcommand{\algcperfprof}{NBIPOP-aCMA}
 \newcommand{\algdperfprof}{NIPOP-aCMA}

\newcommand{\bbobdatapath}{ppdata/}
\providecommand{\bbobppfigsftarget}{\ensuremath{10^{-8}}}
\providecommand{\bbobppfigslegend}[1]{
Expected running time (\ERT\ in number of $f$-evaluations) divided by dimension for target function value \bbobppfigsftarget\ as $\log_{10}$ values versus dimension. Different symbols correspond to different algorithms given in the legend of #1. Light symbols give the maximum number of function evaluations from the longest trial divided by dimension. Horizontal lines give linear scaling, slanted dotted lines give quadratic scaling. Black stars indicate statistically better result compared to all other algorithms with $p<0.01$ and Bonferroni correction number of dimensions (six).  
Legend: 
{\color{blue}$\circ$}: \algorithmA
, {\color{red}$\triangledown$}: \algorithmB
, {\color{cyan}$\star$}: \algorithmC
, {\color{magenta}$\Box$}: \algorithmD
}
\providecommand{\bbobpptablesmanylegend}[1]{Expected running time (ERT in number of function evaluations)
                     divided by the respective best ERT measured during BBOB-2009 (given in 
                     the respective first row) for different $\Df$ values in #1. 
                     The central 80\% range divided by two is given in braces. 
                     The median number of conducted function evaluations is additionally given in 
                     \textit{italics}, if $\ERT(10^{-7}) = \infty$.
                     \#succ is the number of trials that reached the final target $\fopt + 10^{-8}$.
                     Best results are printed in bold. }
\providecommand{\algorithmA}{BIPOP-aCMA}
\providecommand{\algorithmB}{IPOP-aCMA}
\providecommand{\algorithmC}{NBIPOP-aCMA}
\providecommand{\algorithmD}{NIPOP-aCMA}
\providecommand{\algorithmA}{BIPOP-aCMA}
\providecommand{\algorithmB}{IPOP-aCMA}
\providecommand{\algorithmC}{NBIPOP-aCMA}
\providecommand{\algorithmD}{NIPOP-aCMA}
\providecommand{\bbobppfigsftarget}{\ensuremath{10^{-8}}}
\providecommand{\bbobppfigslegend}[1]{
Expected running time (\ERT\ in number of $f$-evaluations) divided by dimension for target function value \bbobppfigsftarget\ as $\log_{10}$ values versus dimension. Different symbols correspond to different algorithms given in the legend of #1. Light symbols give the maximum number of function evaluations from the longest trial divided by dimension. Horizontal lines give linear scaling, slanted dotted lines give quadratic scaling. Black stars indicate statistically better result compared to all other algorithms with $p<0.01$ and Bonferroni correction number of dimensions (six).  
Legend: 
{\color{blue}$\circ$}: \algorithmA
, {\color{red}$\triangledown$}: \algorithmB
, {\color{cyan}$\star$}: \algorithmC
, {\color{magenta}$\Box$}: \algorithmD
}
\providecommand{\bbobpptablesmanylegend}[1]{Expected running time (ERT in number of function evaluations)
                     divided by the respective best ERT measured during BBOB-2009 (given in 
                     the respective first row) for different $\Df$ values in #1. 
                     The central 80\% range divided by two is given in braces. 
                     The median number of conducted function evaluations is additionally given in 
                     \textit{italics}, if $\ERT(10^{-7}) = \infty$.
                     \#succ is the number of trials that reached the final target $\fopt + 10^{-8}$.
                     Best results are printed in bold. }
\providecommand{\algorithmA}{BIPOP-aCMA}
\providecommand{\algorithmB}{IPOP-aCMA}
\providecommand{\algorithmC}{NBIPOP-aCMA}
\providecommand{\algorithmD}{NIPOP-aCMA}
\providecommand{\bbobppfigsftarget}{\ensuremath{10^{-8}}}
\providecommand{\bbobppfigslegend}[1]{
Expected running time (\ERT\ in number of $f$-evaluations) divided by dimension for target function value \bbobppfigsftarget\ as $\log_{10}$ values versus dimension. Different symbols correspond to different algorithms given in the legend of #1. Light symbols give the maximum number of function evaluations from the longest trial divided by dimension. Horizontal lines give linear scaling, slanted dotted lines give quadratic scaling. Black stars indicate statistically better result compared to all other algorithms with $p<0.01$ and Bonferroni correction number of dimensions (six).  
Legend: 
{\color{blue}$\circ$}: \algorithmA
, {\color{red}$\triangledown$}: \algorithmB
, {\color{cyan}$\star$}: \algorithmC
, {\color{magenta}$\Box$}: \algorithmD
}
\providecommand{\bbobpptablesmanylegend}[1]{Expected running time (ERT in number of function evaluations)
                     divided by the respective best ERT measured during BBOB-2009 (given in 
                     the respective first row) for different $\Df$ values in #1. 
                     The central 80\% range divided by two is given in braces. 
                     The median number of conducted function evaluations is additionally given in 
                     \textit{italics}, if $\ERT(10^{-7}) = \infty$.
                     \#succ is the number of trials that reached the final target $\fopt + 10^{-8}$.
                     Best results are printed in bold. }
\providecommand{\algorithmA}{BIPOP-aCMA}
\providecommand{\algorithmB}{IPOP-aCMA}
\providecommand{\algorithmC}{NBIPOP-aCMA}
\providecommand{\algorithmD}{NIPOP-aCMA}

\graphicspath{{\bbobdatapath}}

\newcommand{\rot}[2][2.5]{
  \hspace*{-3.5\baselineskip}%
  \begin{rotate}{90}\hspace{#1em}#2
  \end{rotate}}
\newcommand{\includeperfprof}[1]{
  \input{\bbobdatapath #1}%
  \includegraphics[width=0.4135\textwidth,trim=0mm 0mm 34mm 10mm, clip]{#1}%
  \raisebox{.037\textwidth}{\parbox[b][.3\textwidth]{.0868\textwidth}{\begin{scriptsize}
    \perfprofsidepanel 
  \end{scriptsize}}}
}
\begin{figure*}
 \begin{tabular}{@{}c@{}c@{}}
 weakly structured multi-modal fcts & all functions\\
  \input{\bbobdatapath pprldmany_40D_mult2}%
  \includegraphics[width=0.4135\textwidth,trim=0mm 0mm 34mm 10mm, clip]{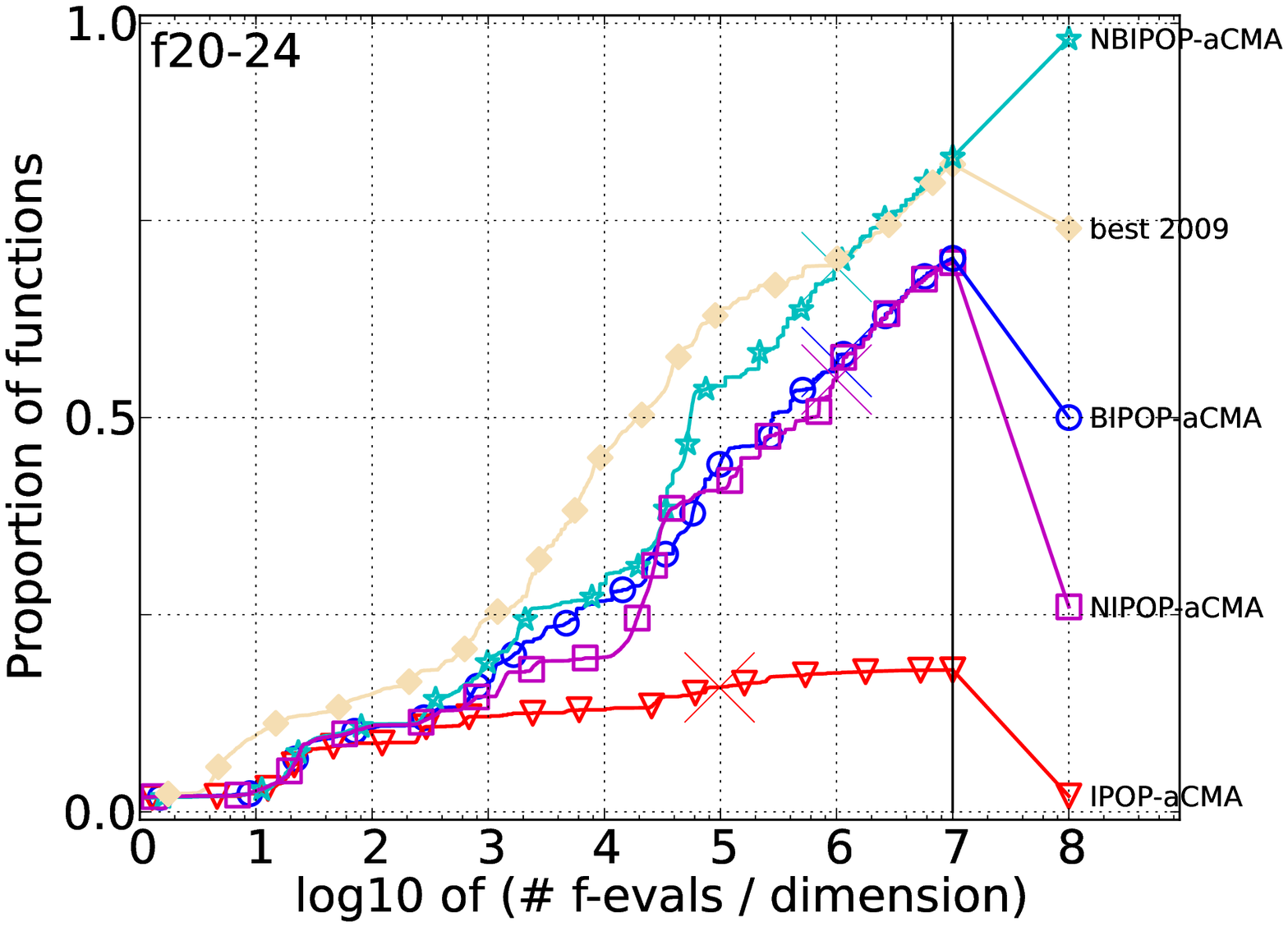}%
  \raisebox{.037\textwidth}{\parbox[b][.3\textwidth]{.0868\textwidth}{\begin{scriptsize}
    \perfprofsidepanel 
  \end{scriptsize}}}
 & 
  \input{\bbobdatapath pprldmany_40D_noiselessall}%
  \includegraphics[width=0.4135\textwidth,trim=0mm 0mm 34mm 10mm, clip]{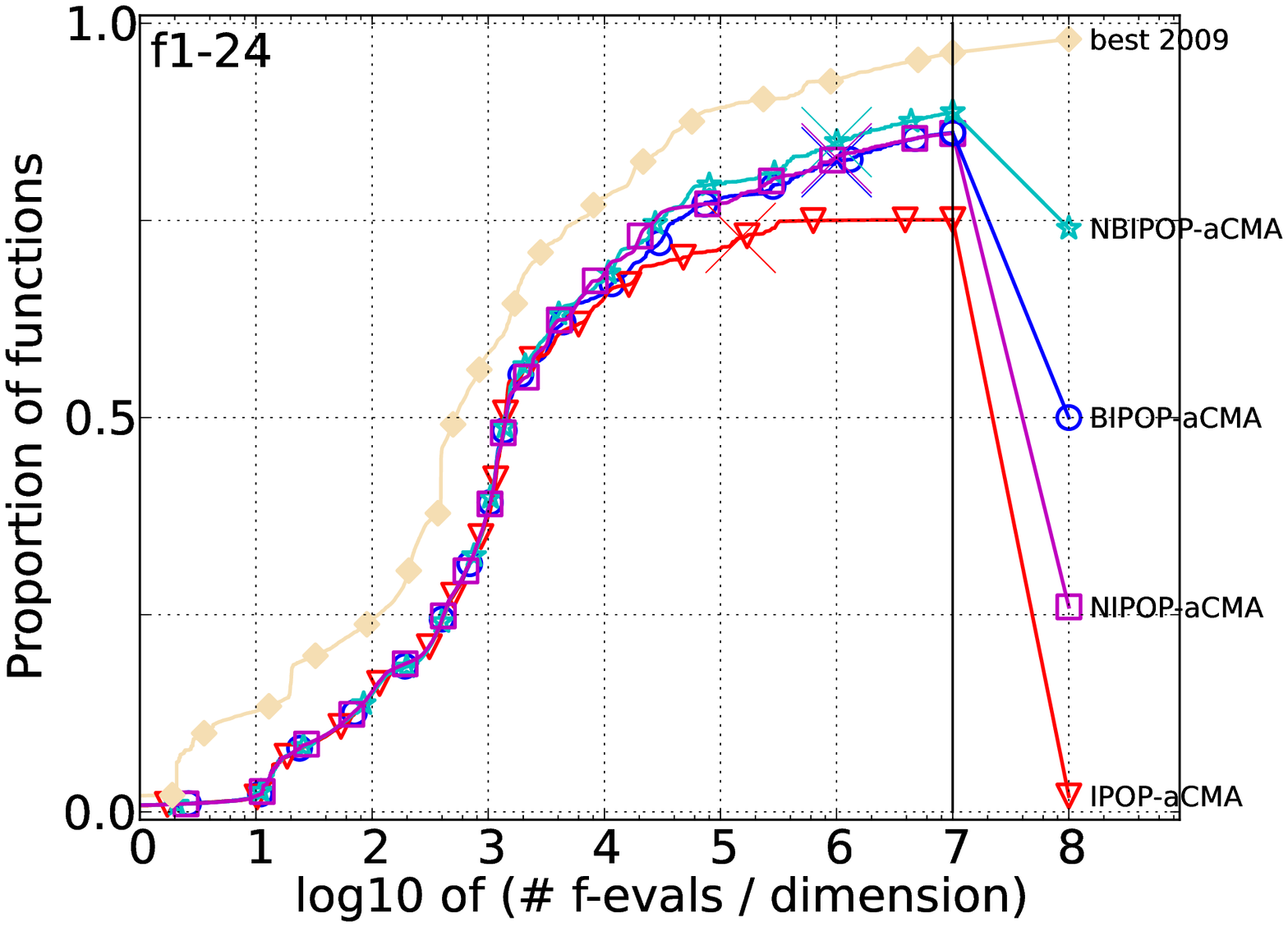}%
  \raisebox{.037\textwidth}{\parbox[b][.3\textwidth]{.0868\textwidth}{\begin{scriptsize}
    \perfprofsidepanel 
  \end{scriptsize}}}
 
 \end{tabular}
\caption{
\label{fig:ECDFs40D}
Bootstrapped empirical cumulative distribution of 
the number of objective function evaluations
divided by dimension (FEvals/D) for 50 targets in
$10^{[-8..2]}$ for all functions and weakly structured multi-modal subgroup in 40-D. The ``best 2009'' line
corresponds to the best \ERT\ observed during BBOB 2009 for each single target. 
}
\end{figure*}

\begin{table*}\tiny
\mbox{\begin{minipage}[t]{0.48\textwidth}\tiny
\centering

\providecommand{\ntables}{7}\providecommand{\algatables}{\StrLeft{BIPOP-aCMA}{\ntables}}\providecommand{\algbtables}{\StrLeft{IPOP-aCMA}{\ntables}}\providecommand{\algctables}{\StrLeft{NBIPOP-aCMA}{\ntables}}\providecommand{\algdtables}{\StrLeft{NIPOP-aCMA}{\ntables}}
\begin{tabularx}{1.03\textwidth}{@{}c@{}|*{6}{@{}r@{}X@{}}|@{}r@{}@{}l@{}}
$\Delta f_\mathrm{opt}$ & \multicolumn{2}{@{\,}X@{\,}}{1e1} & \multicolumn{2}{@{\,}X@{\,}}{1e0} & \multicolumn{2}{@{\,}X@{\,}}{1e-1} & \multicolumn{2}{@{\,}X@{\,}}{1e-3} & \multicolumn{2}{@{\,}X@{\,}}{1e-5} & \multicolumn{2}{@{\,}X@{}|}{1e-7} & \multicolumn{2}{@{}l@{}}{\#succ}\\\hline
\textbf{f3} & \multicolumn{2}{@{\,}X@{\,}}{15526} & \multicolumn{2}{@{\,}X@{\,}}{15602} & \multicolumn{2}{@{\,}X@{\,}}{15612} & \multicolumn{2}{@{\,}X@{\,}}{15646} & \multicolumn{2}{@{\,}X@{\,}}{15651} & \multicolumn{2}{@{\,}X@{\,}|}{15656} & 15 & /15\\
\algatables\hspace*{\fill} & \textbf{2395} & \textbf{} & \multicolumn{2}{@{\,}X@{\,}}{\textbf{$\infty$}} & \multicolumn{2}{@{\,}X@{\,}}{\textbf{$\infty$}} & \multicolumn{2}{@{\,}X@{\,}}{\textbf{$\infty$}} & \multicolumn{2}{@{\,}X@{\,}}{\textbf{$\infty$}} & \multicolumn{2}{@{\,}X@{\,}|}{$\infty$\,\textit{4e7}} & 0 & /15\\
\algbtables\hspace*{\fill} & \multicolumn{2}{@{\,}X@{\,}}{$\infty$} & \multicolumn{2}{@{\,}X@{\,}}{\textbf{$\infty$}} & \multicolumn{2}{@{\,}X@{\,}}{\textbf{$\infty$}} & \multicolumn{2}{@{\,}X@{\,}}{\textbf{$\infty$}} & \multicolumn{2}{@{\,}X@{\,}}{\textbf{$\infty$}} & \multicolumn{2}{@{\,}X@{\,}|}{$\infty$\,\textit{6e6}} & 0 & /8\\
\algctables\hspace*{\fill} & 8177 &  & \multicolumn{2}{@{\,}X@{\,}}{\textbf{$\infty$}} & \multicolumn{2}{@{\,}X@{\,}}{\textbf{$\infty$}} & \multicolumn{2}{@{\,}X@{\,}}{\textbf{$\infty$}} & \multicolumn{2}{@{\,}X@{\,}}{\textbf{$\infty$}} & \multicolumn{2}{@{\,}X@{\,}|}{$\infty$\,\textit{4e7}} & 0 & /15\\
\algdtables\hspace*{\fill} & 4615 &  & \multicolumn{2}{@{\,}X@{\,}}{\textbf{$\infty$}} & \multicolumn{2}{@{\,}X@{\,}}{\textbf{$\infty$}} & \multicolumn{2}{@{\,}X@{\,}}{\textbf{$\infty$}} & \multicolumn{2}{@{\,}X@{\,}}{\textbf{$\infty$}} & \multicolumn{2}{@{\,}X@{\,}|}{$\infty$\,\textit{4e7}} & 0 & /15
\end{tabularx}

\providecommand{\ntables}{7}\providecommand{\algatables}{\StrLeft{BIPOP-aCMA}{\ntables}}\providecommand{\algbtables}{\StrLeft{IPOP-aCMA}{\ntables}}\providecommand{\algctables}{\StrLeft{NBIPOP-aCMA}{\ntables}}\providecommand{\algdtables}{\StrLeft{NIPOP-aCMA}{\ntables}}
\begin{tabularx}{1.03\textwidth}{@{}c@{}|*{6}{@{}r@{}X@{}}|@{}r@{}@{}l@{}}
$\Delta f_\mathrm{opt}$ & \multicolumn{2}{@{\,}X@{\,}}{1e1} & \multicolumn{2}{@{\,}X@{\,}}{1e0} & \multicolumn{2}{@{\,}X@{\,}}{1e-1} & \multicolumn{2}{@{\,}X@{\,}}{1e-3} & \multicolumn{2}{@{\,}X@{\,}}{1e-5} & \multicolumn{2}{@{\,}X@{}|}{1e-7} & \multicolumn{2}{@{}l@{}}{\#succ}\\\hline
\textbf{f4} & \multicolumn{2}{@{\,}X@{\,}}{15536} & \multicolumn{2}{@{\,}X@{\,}}{15601} & \multicolumn{2}{@{\,}X@{\,}}{15659} & \multicolumn{2}{@{\,}X@{\,}}{15703} & \multicolumn{2}{@{\,}X@{\,}}{15733} & \multicolumn{2}{@{\,}X@{\,}|}{2.8e5} & 6 & /15\\
\algatables\hspace*{\fill} & \multicolumn{2}{@{\,}X@{\,}}{\textbf{$\infty$}} & \multicolumn{2}{@{\,}X@{\,}}{\textbf{$\infty$}} & \multicolumn{2}{@{\,}X@{\,}}{\textbf{$\infty$}} & \multicolumn{2}{@{\,}X@{\,}}{\textbf{$\infty$}} & \multicolumn{2}{@{\,}X@{\,}}{\textbf{$\infty$}} & \multicolumn{2}{@{\,}X@{\,}|}{$\infty$\,\textit{4e7}} & 0 & /15\\
\algbtables\hspace*{\fill} & \multicolumn{2}{@{\,}X@{\,}}{\textbf{$\infty$}} & \multicolumn{2}{@{\,}X@{\,}}{\textbf{$\infty$}} & \multicolumn{2}{@{\,}X@{\,}}{\textbf{$\infty$}} & \multicolumn{2}{@{\,}X@{\,}}{\textbf{$\infty$}} & \multicolumn{2}{@{\,}X@{\,}}{\textbf{$\infty$}} & \multicolumn{2}{@{\,}X@{\,}|}{$\infty$\,\textit{6e6}} & 0 & /8\\
\algctables\hspace*{\fill} & \multicolumn{2}{@{\,}X@{\,}}{\textbf{$\infty$}} & \multicolumn{2}{@{\,}X@{\,}}{\textbf{$\infty$}} & \multicolumn{2}{@{\,}X@{\,}}{\textbf{$\infty$}} & \multicolumn{2}{@{\,}X@{\,}}{\textbf{$\infty$}} & \multicolumn{2}{@{\,}X@{\,}}{\textbf{$\infty$}} & \multicolumn{2}{@{\,}X@{\,}|}{$\infty$\,\textit{4e7}} & 0 & /15\\
\algdtables\hspace*{\fill} & \multicolumn{2}{@{\,}X@{\,}}{\textbf{$\infty$}} & \multicolumn{2}{@{\,}X@{\,}}{\textbf{$\infty$}} & \multicolumn{2}{@{\,}X@{\,}}{\textbf{$\infty$}} & \multicolumn{2}{@{\,}X@{\,}}{\textbf{$\infty$}} & \multicolumn{2}{@{\,}X@{\,}}{\textbf{$\infty$}} & \multicolumn{2}{@{\,}X@{\,}|}{$\infty$\,\textit{4e7}} & 0 & /15
\end{tabularx}

\providecommand{\ntables}{7}\providecommand{\algatables}{\StrLeft{BIPOP-aCMA}{\ntables}}\providecommand{\algbtables}{\StrLeft{IPOP-aCMA}{\ntables}}\providecommand{\algctables}{\StrLeft{NBIPOP-aCMA}{\ntables}}\providecommand{\algdtables}{\StrLeft{NIPOP-aCMA}{\ntables}}
\begin{tabularx}{1.03\textwidth}{@{}c@{}|*{6}{@{}r@{}X@{}}|@{}r@{}@{}l@{}}
$\Delta f_\mathrm{opt}$ & \multicolumn{2}{@{\,}X@{\,}}{1e1} & \multicolumn{2}{@{\,}X@{\,}}{1e0} & \multicolumn{2}{@{\,}X@{\,}}{1e-1} & \multicolumn{2}{@{\,}X@{\,}}{1e-3} & \multicolumn{2}{@{\,}X@{\,}}{1e-5} & \multicolumn{2}{@{\,}X@{}|}{1e-7} & \multicolumn{2}{@{}l@{}}{\#succ}\\\hline
\textbf{f15} & \multicolumn{2}{@{\,}X@{\,}}{1.9e5} & \multicolumn{2}{@{\,}X@{\,}}{7.9e5} & \multicolumn{2}{@{\,}X@{\,}}{1.0e6} & \multicolumn{2}{@{\,}X@{\,}}{1.1e6} & \multicolumn{2}{@{\,}X@{\,}}{1.1e6} & \multicolumn{2}{@{\,}X@{\,}|}{1.1e6} & 15 & /15\\
\algatables\hspace*{\fill} & 1 & .2 & 1 & .1 & 1 & .1 & 1 & .1 & 1 & .1 & 1 & .1 & 15 & /15\\
\algbtables\hspace*{\fill} & \textbf{0} & \textbf{.72} & \textbf{0} & \textbf{.43} & 0 & .60 & 0 & .61 & 0 & .62 & 0 & .63 & 8 & /8\\
\algctables\hspace*{\fill} & 1 & .0 & 0 & .71 & 0 & .75 & 0 & .76 & 0 & .77 & 0 & .77 & 15 & /15\\
\algdtables\hspace*{\fill} & 0 & .92 & 0 & .61 & \textbf{0} & \textbf{.55} & \textbf{0} & \textbf{.56} & \textbf{0} & \textbf{.57} & \textbf{0} & \textbf{.58} & 15 & /15
\end{tabularx}

\providecommand{\ntables}{7}\providecommand{\algatables}{\StrLeft{BIPOP-aCMA}{\ntables}}\providecommand{\algbtables}{\StrLeft{IPOP-aCMA}{\ntables}}\providecommand{\algctables}{\StrLeft{NBIPOP-aCMA}{\ntables}}\providecommand{\algdtables}{\StrLeft{NIPOP-aCMA}{\ntables}}
\begin{tabularx}{1.03\textwidth}{@{}c@{}|*{6}{@{}r@{}X@{}}|@{}r@{}@{}l@{}}
$\Delta f_\mathrm{opt}$ & \multicolumn{2}{@{\,}X@{\,}}{1e1} & \multicolumn{2}{@{\,}X@{\,}}{1e0} & \multicolumn{2}{@{\,}X@{\,}}{1e-1} & \multicolumn{2}{@{\,}X@{\,}}{1e-3} & \multicolumn{2}{@{\,}X@{\,}}{1e-5} & \multicolumn{2}{@{\,}X@{}|}{1e-7} & \multicolumn{2}{@{}l@{}}{\#succ}\\\hline
\textbf{f16} & \multicolumn{2}{@{\,}X@{\,}}{5244} & \multicolumn{2}{@{\,}X@{\,}}{72122} & \multicolumn{2}{@{\,}X@{\,}}{3.2e5} & \multicolumn{2}{@{\,}X@{\,}}{1.4e6} & \multicolumn{2}{@{\,}X@{\,}}{2.0e6} & \multicolumn{2}{@{\,}X@{\,}|}{2.0e6} & 15 & /15\\
\algatables\hspace*{\fill} & 1 & .3 & 0 & .96 & 0 & .80 & 0 & .54 & 0 & .50 & 0 & .51 & 15 & /15\\
\algbtables\hspace*{\fill} & \textbf{0} & \textbf{.91} & 1 & .1 & 1 & .0 & 0 & .51 & 1 & .4 & 1 & .4 & 8 & /8\\
\algctables\hspace*{\fill} & 0 & .97 & 0 & .78 & 0 & .34 & 0 & .38 & 0 & .46 & 0 & .74 & 15 & /15\\
\algdtables\hspace*{\fill} & 1 & .2 & \textbf{0} & \textbf{.65} & \textbf{0} & \textbf{.23} & \textbf{0} & \textbf{.21} & \textbf{0} & \textbf{.16} & \textbf{0} & \textbf{.18} & 15 & /15
\end{tabularx}

\providecommand{\ntables}{7}\providecommand{\algatables}{\StrLeft{BIPOP-aCMA}{\ntables}}\providecommand{\algbtables}{\StrLeft{IPOP-aCMA}{\ntables}}\providecommand{\algctables}{\StrLeft{NBIPOP-aCMA}{\ntables}}\providecommand{\algdtables}{\StrLeft{NIPOP-aCMA}{\ntables}}
\begin{tabularx}{1.03\textwidth}{@{}c@{}|*{6}{@{}r@{}X@{}}|@{}r@{}@{}l@{}}
$\Delta f_\mathrm{opt}$ & \multicolumn{2}{@{\,}X@{\,}}{1e1} & \multicolumn{2}{@{\,}X@{\,}}{1e0} & \multicolumn{2}{@{\,}X@{\,}}{1e-1} & \multicolumn{2}{@{\,}X@{\,}}{1e-3} & \multicolumn{2}{@{\,}X@{\,}}{1e-5} & \multicolumn{2}{@{\,}X@{}|}{1e-7} & \multicolumn{2}{@{}l@{}}{\#succ}\\\hline
\textbf{f17} & \multicolumn{2}{@{\,}X@{\,}}{399} & \multicolumn{2}{@{\,}X@{\,}}{4220} & \multicolumn{2}{@{\,}X@{\,}}{14158} & \multicolumn{2}{@{\,}X@{\,}}{51958} & \multicolumn{2}{@{\,}X@{\,}}{1.3e5} & \multicolumn{2}{@{\,}X@{\,}|}{2.7e5} & 14 & /15\\
\algatables\hspace*{\fill} & 1 & .1 & 0 & .64 & 1 & .6 & 1 & .1 & 1 & .4 & 0 & .87 & 15 & /15\\
\algbtables\hspace*{\fill} & 1 & .0 & 0 & .52 & 1 & .3 & 1 & .3 & \textbf{0} & \textbf{.97} & 0 & .83 & 8 & /8\\
\algctables\hspace*{\fill} & 1 & .0 & 0 & .57 & 1 & .2 & 1 & .2 & 1 & .0 & 0 & .81 & 15 & /15\\
\algdtables\hspace*{\fill} & \textbf{0} & \textbf{.97} & \textbf{0} & \textbf{.52} & \textbf{0} & \textbf{.97} & \textbf{1} & \textbf{.00} & 1 & .1 & \textbf{0} & \textbf{.70} & 15 & /15
\end{tabularx}

\providecommand{\ntables}{7}\providecommand{\algatables}{\StrLeft{BIPOP-aCMA}{\ntables}}\providecommand{\algbtables}{\StrLeft{IPOP-aCMA}{\ntables}}\providecommand{\algctables}{\StrLeft{NBIPOP-aCMA}{\ntables}}\providecommand{\algdtables}{\StrLeft{NIPOP-aCMA}{\ntables}}
\begin{tabularx}{1.03\textwidth}{@{}c@{}|*{6}{@{}r@{}X@{}}|@{}r@{}@{}l@{}}
$\Delta f_\mathrm{opt}$ & \multicolumn{2}{@{\,}X@{\,}}{1e1} & \multicolumn{2}{@{\,}X@{\,}}{1e0} & \multicolumn{2}{@{\,}X@{\,}}{1e-1} & \multicolumn{2}{@{\,}X@{\,}}{1e-3} & \multicolumn{2}{@{\,}X@{\,}}{1e-5} & \multicolumn{2}{@{\,}X@{}|}{1e-7} & \multicolumn{2}{@{}l@{}}{\#succ}\\\hline
\textbf{f18} & \multicolumn{2}{@{\,}X@{\,}}{1442} & \multicolumn{2}{@{\,}X@{\,}}{16998} & \multicolumn{2}{@{\,}X@{\,}}{47068} & \multicolumn{2}{@{\,}X@{\,}}{1.9e5} & \multicolumn{2}{@{\,}X@{\,}}{6.7e5} & \multicolumn{2}{@{\,}X@{\,}|}{9.5e5} & 15 & /15\\
\algatables\hspace*{\fill} & \textbf{0} & \textbf{.94} & \textbf{0} & \textbf{.51} & 1 & .0 & 0 & .98 & 0 & .88 & 0 & .67 & 15 & /15\\
\algbtables\hspace*{\fill} & 0 & .96 & 0 & .68 & 1 & .0 & \textbf{0} & \textbf{.66} & \textbf{0} & \textbf{.45} & 0 & .48 & 8 & /8\\
\algctables\hspace*{\fill} & 1 & .0 & 0 & .97 & 1 & .1 & 0 & .93 & 0 & .57 & 0 & .53 & 15 & /15\\
\algdtables\hspace*{\fill} & 0 & .95 & 0 & .58 & \textbf{0} & \textbf{.75} & 0 & .71 & 0 & .50 & \textbf{0} & \textbf{.42} & 15 & /15
\end{tabularx}

\end{minipage}

\hspace{3mm}

\begin{minipage}[t]{0.48\textwidth}\tiny
\centering

\providecommand{\ntables}{7}\providecommand{\algatables}{\StrLeft{BIPOP-aCMA}{\ntables}}\providecommand{\algbtables}{\StrLeft{IPOP-aCMA}{\ntables}}\providecommand{\algctables}{\StrLeft{NBIPOP-aCMA}{\ntables}}\providecommand{\algdtables}{\StrLeft{NIPOP-aCMA}{\ntables}}
\begin{tabularx}{1.03\textwidth}{@{}c@{}|*{6}{@{}r@{}X@{}}|@{}r@{}@{}l@{}}
$\Delta f_\mathrm{opt}$ & \multicolumn{2}{@{\,}X@{\,}}{1e1} & \multicolumn{2}{@{\,}X@{\,}}{1e0} & \multicolumn{2}{@{\,}X@{\,}}{1e-1} & \multicolumn{2}{@{\,}X@{\,}}{1e-3} & \multicolumn{2}{@{\,}X@{\,}}{1e-5} & \multicolumn{2}{@{\,}X@{}|}{1e-7} & \multicolumn{2}{@{}l@{}}{\#succ}\\\hline
\textbf{f19} & \multicolumn{2}{@{\,}X@{\,}}{1} & \multicolumn{2}{@{\,}X@{\,}}{1} & \multicolumn{2}{@{\,}X@{\,}}{1.4e6} & \multicolumn{2}{@{\,}X@{\,}}{2.6e7} & \multicolumn{2}{@{\,}X@{\,}}{4.5e7} & \multicolumn{2}{@{\,}X@{\,}|}{4.5e7} & 8 & /15\\
\algatables\hspace*{\fill} & \textbf{396} & \textbf{} & \multicolumn{2}{@{\,}X@{\,}}{6.7e4} & 0 & .87 & 1 & .2 & 1 & .0 & 1 & .0 & 9 & /15\\
\algbtables\hspace*{\fill} & 462 &  & \multicolumn{2}{@{\,}X@{\,}}{\textbf{4.4e4}} & \textbf{0} & \textbf{.57} & \textbf{0} & \textbf{.34} & \textbf{0} & \textbf{.20} & \textbf{0} & \textbf{.20} & 8 & /8\\
\algctables\hspace*{\fill} & 424 &  & \multicolumn{2}{@{\,}X@{\,}}{8.3e4} & 0 & .97 & 0 & .81 & 1 & .1 & 1 & .1 & 9 & /15\\
\algdtables\hspace*{\fill} & 436 &  & \multicolumn{2}{@{\,}X@{\,}}{8.2e4} & 1 & .9 & 0 & .48 & 0 & .32 & 0 & .32 & 15 & /15
\end{tabularx}

\providecommand{\ntables}{7}\providecommand{\algatables}{\StrLeft{BIPOP-aCMA}{\ntables}}\providecommand{\algbtables}{\StrLeft{IPOP-aCMA}{\ntables}}\providecommand{\algctables}{\StrLeft{NBIPOP-aCMA}{\ntables}}\providecommand{\algdtables}{\StrLeft{NIPOP-aCMA}{\ntables}}
\begin{tabularx}{1.03\textwidth}{@{}c@{}|*{6}{@{}r@{}X@{}}|@{}r@{}@{}l@{}}
$\Delta f_\mathrm{opt}$ & \multicolumn{2}{@{\,}X@{\,}}{1e1} & \multicolumn{2}{@{\,}X@{\,}}{1e0} & \multicolumn{2}{@{\,}X@{\,}}{1e-1} & \multicolumn{2}{@{\,}X@{\,}}{1e-3} & \multicolumn{2}{@{\,}X@{\,}}{1e-5} & \multicolumn{2}{@{\,}X@{}|}{1e-7} & \multicolumn{2}{@{}l@{}}{\#succ}\\\hline
\textbf{f20} & \multicolumn{2}{@{\,}X@{\,}}{222} & \multicolumn{2}{@{\,}X@{\,}}{1.3e5} & \multicolumn{2}{@{\,}X@{\,}}{1.6e8} & \multicolumn{2}{@{\,}X@{\,}}{$\infty$} & \multicolumn{2}{@{\,}X@{\,}}{$\infty$} & \multicolumn{2}{@{\,}X@{\,}|}{$\infty$} & 0\\
\algatables\hspace*{\fill} & 4 & .0 & 9 & .0 & 0 & .34 & \multicolumn{2}{@{\,}X@{\,}}{.} & \multicolumn{2}{@{\,}X@{\,}}{.} & \multicolumn{2}{@{\,}X@{\,}|}{.} & 0 & /15\\
\algbtables\hspace*{\fill} & \textbf{3} & \textbf{.9} & 8 & .1 & \textbf{0} & \textbf{.18} & \multicolumn{2}{@{\,}X@{\,}}{.} & \multicolumn{2}{@{\,}X@{\,}}{.} & \multicolumn{2}{@{\,}X@{\,}|}{.} & 0 & /8\\
\algctables\hspace*{\fill} & 4 & .0 & 8 & .5 & 0 & .39 & \multicolumn{2}{@{\,}X@{\,}}{.} & \multicolumn{2}{@{\,}X@{\,}}{.} & \multicolumn{2}{@{\,}X@{\,}|}{.} & 0 & /15\\
\algdtables\hspace*{\fill} & 4 & .0 & \textbf{6} & \textbf{.5} & 0 & .32 & \multicolumn{2}{@{\,}X@{\,}}{.} & \multicolumn{2}{@{\,}X@{\,}}{.} & \multicolumn{2}{@{\,}X@{\,}|}{.} & 0 & /15
\end{tabularx}

\providecommand{\ntables}{7}\providecommand{\algatables}{\StrLeft{BIPOP-aCMA}{\ntables}}\providecommand{\algbtables}{\StrLeft{IPOP-aCMA}{\ntables}}\providecommand{\algctables}{\StrLeft{NBIPOP-aCMA}{\ntables}}\providecommand{\algdtables}{\StrLeft{NIPOP-aCMA}{\ntables}}
\begin{tabularx}{1.03\textwidth}{@{}c@{}|*{6}{@{}r@{}X@{}}|@{}r@{}@{}l@{}}
$\Delta f_\mathrm{opt}$ & \multicolumn{2}{@{\,}X@{\,}}{1e1} & \multicolumn{2}{@{\,}X@{\,}}{1e0} & \multicolumn{2}{@{\,}X@{\,}}{1e-1} & \multicolumn{2}{@{\,}X@{\,}}{1e-3} & \multicolumn{2}{@{\,}X@{\,}}{1e-5} & \multicolumn{2}{@{\,}X@{}|}{1e-7} & \multicolumn{2}{@{}l@{}}{\#succ}\\\hline
\textbf{f21} & \multicolumn{2}{@{\,}X@{\,}}{1044} & \multicolumn{2}{@{\,}X@{\,}}{21144} & \multicolumn{2}{@{\,}X@{\,}}{1.0e5} & \multicolumn{2}{@{\,}X@{\,}}{1.0e5} & \multicolumn{2}{@{\,}X@{\,}}{1.0e5} & \multicolumn{2}{@{\,}X@{\,}|}{1.0e5} & 26 & /30\\
\algatables\hspace*{\fill} & 7 & .5 & 60 &  & 37 &  & 37 &  & 37 &  & 37 &  & 15 & /15\\
\algbtables\hspace*{\fill} & 7 & .1 & 421 &  & \multicolumn{2}{@{\,}X@{\,}}{$\infty$} & \multicolumn{2}{@{\,}X@{\,}}{$\infty$} & \multicolumn{2}{@{\,}X@{\,}}{$\infty$} & \multicolumn{2}{@{\,}X@{\,}|}{$\infty$\,\textit{3e6}} & 0 & /8\\
\algctables\hspace*{\fill} & \textbf{4} & \textbf{.9} & \textbf{10} & \textbf{} & \textbf{5} & \textbf{.1} & \textbf{5} & \textbf{.1} & \textbf{5} & \textbf{.1} & \textbf{5} & \textbf{.1} & 15 & /15\\
\algdtables\hspace*{\fill} & 14 &  & 440 &  & 173 &  & 172 &  & 171 &  & 171 &  & 12 & /15
\end{tabularx}

\providecommand{\ntables}{7}\providecommand{\algatables}{\StrLeft{BIPOP-aCMA}{\ntables}}\providecommand{\algbtables}{\StrLeft{IPOP-aCMA}{\ntables}}\providecommand{\algctables}{\StrLeft{NBIPOP-aCMA}{\ntables}}\providecommand{\algdtables}{\StrLeft{NIPOP-aCMA}{\ntables}}
\begin{tabularx}{1.03\textwidth}{@{}c@{}|*{6}{@{}r@{}X@{}}|@{}r@{}@{}l@{}}
$\Delta f_\mathrm{opt}$ & \multicolumn{2}{@{\,}X@{\,}}{1e1} & \multicolumn{2}{@{\,}X@{\,}}{1e0} & \multicolumn{2}{@{\,}X@{\,}}{1e-1} & \multicolumn{2}{@{\,}X@{\,}}{1e-3} & \multicolumn{2}{@{\,}X@{\,}}{1e-5} & \multicolumn{2}{@{\,}X@{}|}{1e-7} & \multicolumn{2}{@{}l@{}}{\#succ}\\\hline
\textbf{f22} & \multicolumn{2}{@{\,}X@{\,}}{3090} & \multicolumn{2}{@{\,}X@{\,}}{35442} & \multicolumn{2}{@{\,}X@{\,}}{6.5e5} & \multicolumn{2}{@{\,}X@{\,}}{6.5e5} & \multicolumn{2}{@{\,}X@{\,}}{6.5e5} & \multicolumn{2}{@{\,}X@{\,}|}{6.5e5} & 8 & /30\\
\algatables\hspace*{\fill} & \textbf{12} & \textbf{} & 343 &  & 201 &  & 200 &  & 200 &  & 199 &  & 4 & /15\\
\algbtables\hspace*{\fill} & 144 &  & \textbf{93} & \textbf{} & \multicolumn{2}{@{\,}X@{\,}}{$\infty$} & \multicolumn{2}{@{\,}X@{\,}}{$\infty$} & \multicolumn{2}{@{\,}X@{\,}}{$\infty$} & \multicolumn{2}{@{\,}X@{\,}|}{$\infty$\,\textit{3e6}} & 0 & /8\\
\algctables\hspace*{\fill} & 12 &  & 112 &  & \textbf{32} & \textbf{} & \textbf{32} & \textbf{} & \textbf{32} & \textbf{} & \textbf{32} & \textbf{} & 12 & /15\\
\algdtables\hspace*{\fill} & 179 &  & 583 &  & \multicolumn{2}{@{\,}X@{\,}}{$\infty$} & \multicolumn{2}{@{\,}X@{\,}}{$\infty$} & \multicolumn{2}{@{\,}X@{\,}}{$\infty$} & \multicolumn{2}{@{\,}X@{\,}|}{$\infty$\,\textit{4e7}} & 0 & /15
\end{tabularx}

\providecommand{\ntables}{7}\providecommand{\algatables}{\StrLeft{BIPOP-aCMA}{\ntables}}\providecommand{\algbtables}{\StrLeft{IPOP-aCMA}{\ntables}}\providecommand{\algctables}{\StrLeft{NBIPOP-aCMA}{\ntables}}\providecommand{\algdtables}{\StrLeft{NIPOP-aCMA}{\ntables}}
\begin{tabularx}{1.03\textwidth}{@{}c@{}|*{6}{@{}r@{}X@{}}|@{}r@{}@{}l@{}}
$\Delta f_\mathrm{opt}$ & \multicolumn{2}{@{\,}X@{\,}}{1e1} & \multicolumn{2}{@{\,}X@{\,}}{1e0} & \multicolumn{2}{@{\,}X@{\,}}{1e-1} & \multicolumn{2}{@{\,}X@{\,}}{1e-3} & \multicolumn{2}{@{\,}X@{\,}}{1e-5} & \multicolumn{2}{@{\,}X@{}|}{1e-7} & \multicolumn{2}{@{}l@{}}{\#succ}\\\hline
\textbf{f23} & \multicolumn{2}{@{\,}X@{\,}}{7.1} & \multicolumn{2}{@{\,}X@{\,}}{11925} & \multicolumn{2}{@{\,}X@{\,}}{75453} & \multicolumn{2}{@{\,}X@{\,}}{1.3e6} & \multicolumn{2}{@{\,}X@{\,}}{3.2e6} & \multicolumn{2}{@{\,}X@{\,}|}{3.4e6} & 15 & /15\\
\algatables\hspace*{\fill} & 8 & .4 & \textbf{7} & \textbf{.8} & \textbf{1} & \textbf{.3} & 1 & .9 & 1 & .00 & 0 & .99 & 15 & /15\\
\algbtables\hspace*{\fill} & 9 & .2 & \multicolumn{2}{@{\,}X@{\,}}{$\infty$} & \multicolumn{2}{@{\,}X@{\,}}{$\infty$} & \multicolumn{2}{@{\,}X@{\,}}{$\infty$} & \multicolumn{2}{@{\,}X@{\,}}{$\infty$} & \multicolumn{2}{@{\,}X@{\,}|}{$\infty$\,\textit{4e6}} & 0 & /8\\
\algctables\hspace*{\fill} & 8 & .6 & 10 &  & 1 & .6 & 1 & .3 & 0 & .58 & 0 & .59 & 15 & /15\\
\algdtables\hspace*{\fill} & \textbf{5} & \textbf{.9} & 61 &  & 11 &  & \textbf{0} & \textbf{.72} & \textbf{0} & \textbf{.36} & \textbf{0} & \textbf{.38} & 15 & /15
\end{tabularx}

\providecommand{\ntables}{7}\providecommand{\algatables}{\StrLeft{BIPOP-aCMA}{\ntables}}\providecommand{\algbtables}{\StrLeft{IPOP-aCMA}{\ntables}}\providecommand{\algctables}{\StrLeft{NBIPOP-aCMA}{\ntables}}\providecommand{\algdtables}{\StrLeft{NIPOP-aCMA}{\ntables}}
\begin{tabularx}{1.03\textwidth}{@{}c@{}|*{6}{@{}r@{}X@{}}|@{}r@{}@{}l@{}}
$\Delta f_\mathrm{opt}$ & \multicolumn{2}{@{\,}X@{\,}}{1e1} & \multicolumn{2}{@{\,}X@{\,}}{1e0} & \multicolumn{2}{@{\,}X@{\,}}{1e-1} & \multicolumn{2}{@{\,}X@{\,}}{1e-3} & \multicolumn{2}{@{\,}X@{\,}}{1e-5} & \multicolumn{2}{@{\,}X@{}|}{1e-7} & \multicolumn{2}{@{}l@{}}{\#succ}\\\hline
\textbf{f24} & \multicolumn{2}{@{\,}X@{\,}}{5.8e6} & \multicolumn{2}{@{\,}X@{\,}}{9.8e7} & \multicolumn{2}{@{\,}X@{\,}}{3.0e8} & \multicolumn{2}{@{\,}X@{\,}}{3.0e8} & \multicolumn{2}{@{\,}X@{\,}}{3.0e8} & \multicolumn{2}{@{\,}X@{\,}|}{3.0e8} & 1 & /15\\
\algatables\hspace*{\fill} & 3 & .6 & 1 & .4 & \multicolumn{2}{@{\,}X@{\,}}{$\infty$} & \multicolumn{2}{@{\,}X@{\,}}{$\infty$} & \multicolumn{2}{@{\,}X@{\,}}{$\infty$} & \multicolumn{2}{@{\,}X@{\,}|}{$\infty$\,\textit{4e7}} & 0 & /15\\
\algbtables\hspace*{\fill} & \multicolumn{2}{@{\,}X@{\,}}{$\infty$} & \multicolumn{2}{@{\,}X@{\,}}{$\infty$} & \multicolumn{2}{@{\,}X@{\,}}{$\infty$} & \multicolumn{2}{@{\,}X@{\,}}{$\infty$} & \multicolumn{2}{@{\,}X@{\,}}{$\infty$} & \multicolumn{2}{@{\,}X@{\,}|}{$\infty$\,\textit{1e7}} & 0 & /8\\
\algctables\hspace*{\fill} & 2 & .1 & 0 & .19 & 0 & .97 & 0 & .97 & 0 & .97 & 0 & .97 & 2 & /15\\
\algdtables\hspace*{\fill} & \textbf{1} & \textbf{.2} & \textbf{0} & \textbf{.15} & \textbf{0} & \textbf{.44} & \textbf{0} & \textbf{.44} & \textbf{0} & \textbf{.44} & \textbf{0} & \textbf{.44} & 4 & /15
\end{tabularx}
\end{minipage}} \small
 \caption{\label{tab:ERTs40} 
Overall results on multi-modal functions $f3-4$ and $f15-24$ in dimension $d=40$: Expected running time (ERT in number of function evaluations) divided by the respective best ERT measured during BBOB-2009 for precision $\Df$ ranging in $10^i$, $i = 1\ldots-7$. 
                     The median number of conducted function evaluations is additionally given in 
                     \textit{italics}, if $\ERT(10^{-7}) = \infty$.
                     \#succ is the number of trials that reached the final target $\fopt + 10^{-8}$.
                     Best results are printed in bold. For a more detailed (statistical) analysis of results on BBOB problems, please see \cite{BBOB2012NIPOP}.
										Statistically significantly better entries (Wilcoxon rank-sum test with $p=0.05$) are indicated in bold. The interested reader is
referred to \cite{BBOB2012NIPOP} for the statistical analysis and discussion of these results.
}
\end{table*}

\end{document}